\documentclass[10pt,twocolumn,letterpaper]{article}

\usepackage[pagenumbers]{cvpr} 

\usepackage{graphicx}
\usepackage{amsmath}
\usepackage{amssymb}
\usepackage{booktabs}
\usepackage{bm}

\usepackage{amsthm,amsmath,amssymb}
\usepackage{mathrsfs}
\usepackage[pagebackref,breaklinks,colorlinks]{hyperref}

\usepackage[accsupp]{axessibility}  

\usepackage[capitalize]{cleveref}
\crefname{section}{Sec.}{Secs.}
\Crefname{section}{Section}{Sections}
\Crefname{table}{Table}{Tables}
\crefname{table}{Tab.}{Tabs.}

\def\cvprPaperID{11} 
\def\confName{MULA WORKSHOP}
\def\confYear{2022}

\begin{document}

\title{Reasoning with Multi-Structure Commonsense Knowledge \\ in Visual Dialog}

\author{Shunyu Zhang$^{1}$ \and 
    Xiaoze Jiang$^{1}$ \and
    Zequn Yang$^{1}$ \and
    Tao Wan$^{2}$ \and
    Zengchang Qin$^{1,*}$ \and
$^1$ Intelligent Computing and Machine Learning Lab, School of ASEE, Beihang University\\
$^2$ School of BSME, Beijing Advanced Innovation Center for Biomedical Engineering, Beihang University \\
{\tt\small \{zhangshunyu, xzjiang, zqyang, taowan, $^{*}$zcqin\}@buaa.edu.cn}
}


\maketitle

\begin{abstract}
Visual Dialog requires an agent to engage in a conversation with humans grounded in an image. Many studies on Visual Dialog focus on the understanding of the dialog history or the content of an image, while a considerable amount of commonsense-required questions are ignored. Handling these scenarios depends on logical reasoning that requires commonsense priors. How to capture relevant commonsense knowledge complementary to the history and the image remains a key challenge. In this paper, we propose a novel model by \textbf{R}easoning with \textbf{M}ulti-structure Commonsense \textbf{K}nowledge (RMK). In our model, the external knowledge is represented with sentence-level facts and graph-level facts, to properly suit the scenario of the composite of dialog history and image. On top of these multi-structure representations, our model can capture relevant knowledge and incorporate them into the vision and semantic features, via graph-based interaction and transformer-based fusion. Experimental results and analysis on VisDial v1.0 and VisDialCK datasets show that our proposed model effectively outperforms comparative methods.

\end{abstract}


\section{Introduction}
\label{intro}
\begin{figure}[t]
\centering
\includegraphics[width=1.0\columnwidth]{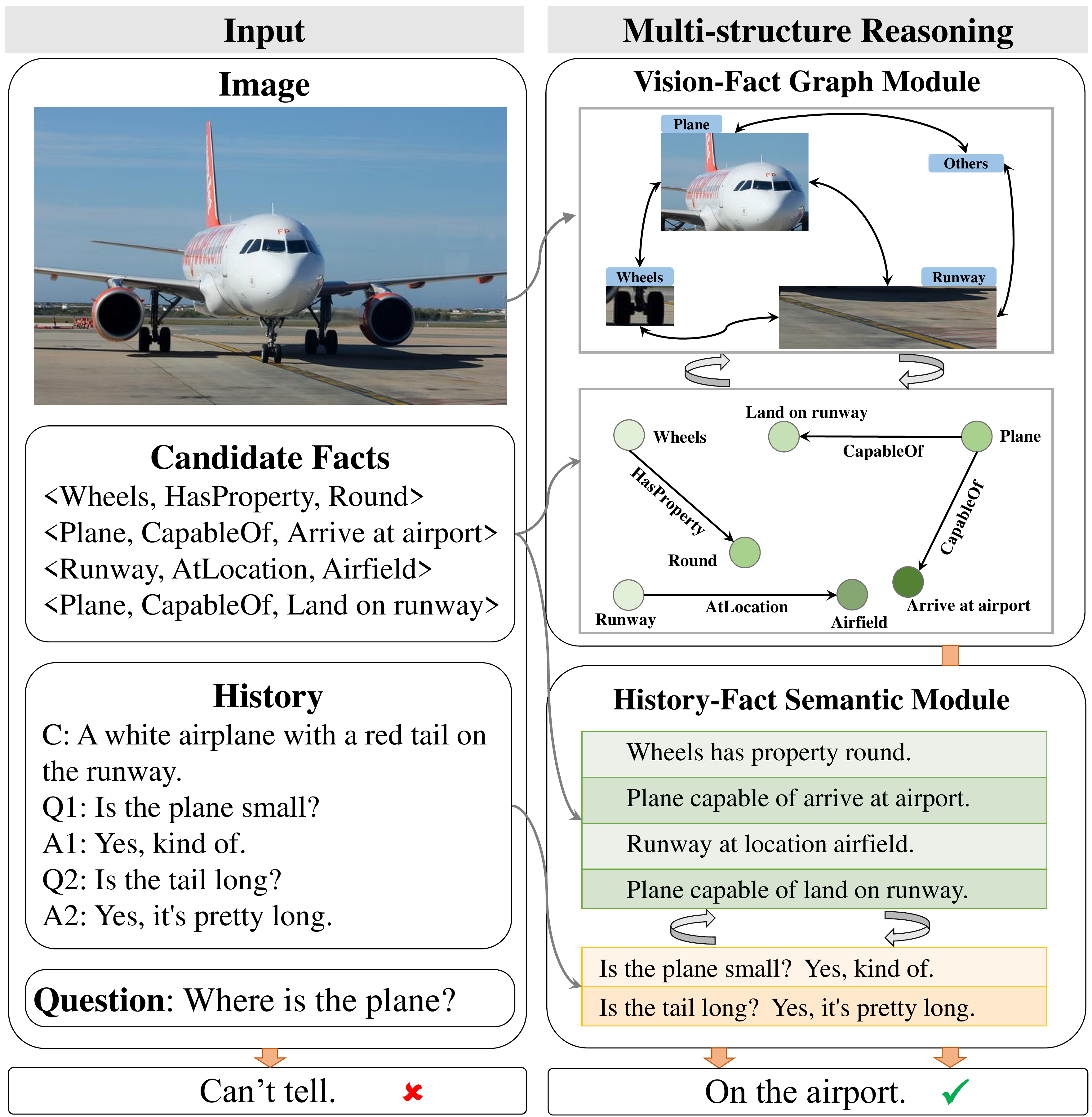}
\caption{An illustration of RMK method. We represent Candidate Facts with multiple structures (graph-level and sentence-level) to reason with image and history for the optimal answer. }
\label{img:idea}
\end{figure}

With the increasing interest in Visual Dialog task \cite{das2017visual}, which involves an agent to make a dialog conditional on an image, there exist loads of studies \cite{Zheng2019Reasoning,Niu2018Recursive,chen-etal-2021-gog} concentrating on the reasoning of dialog history. Some recent works~\cite{agarwal2020history} showed that 10.96$\%$ of the questions in the validation set of the well-known dataset of {\em VisDial v1.0} \cite{das2017visual}  demand dialog history, while there are 10.62$\%$ of questions that require commonsense knowledge from their annotated data. However, there was little research studying the commonsense-required questions, compared to history-required ones. As shown in Figure \ref{img:idea}, when answering ``\textit{Where is the plane?}'', without commonsense knowledge, the agent cannot easily figure out the place where the plane parks and only replies with the safe response ``\textit{Can't tell.}''. Therefore, how to equip a visual dialog system with commonsense knowledge is unresolved and remains a challenge in the Vision and Language research. 

There were quite a few attempts on knowledge-based visual question answering (KB-VQA) \cite{wang2017fvqa,marino2019ok}. The advanced solutions usually build a fact graph with filtered fact triplets and then reason on the graph to infer the best answer \cite{zhu2020mucko,narasimhan2018out}. 
However, Visual Dialog task requires an agent to comprehend the dialog history information additionally compared to the VQA tasks \cite{Agrawal2017VQA}, so calls for more contextual logic. What's more, graph-style knowledge has limited ability in capturing semantic-level information, since it pays more attention to the relationship of the knowledge entities. Thus, the single-structure knowledge at semantic-level or graph-level may not satisfy the unique requirements of the visual dialog tasks.

To solve the above problems, we propose a novel multi-structure knowledge representations: i.e. \textit{graph-level facts} and \textit{sentence-level facts}, incorporating with two essential visual dialog components (i.e. image and dialog history). The graph-level facts are used to model relations in commonsense knowledge, and they can also complement the underlying visual relationship explicitly. Therefore we build a visual graph combined with graph-level facts, as shown in Fig.\ref{img:idea}. On the other side, the sentence-level facts tackle the knowledge semantics, it maps the knowledge in triplet to the text space. We equip them with sentence-level facts to better extract semantic features, for dialog history also contains semantic relations implicitly. Meanwhile, the advantage of this combination is that the image and dialog history is associated with homologous knowledge information, bridging the heterogeneous gap and complementary to each other. 

\begin{figure*}[t]
\setlength{\abovecaptionskip}{2mm}
\centering
\includegraphics[width=1.01\textwidth]{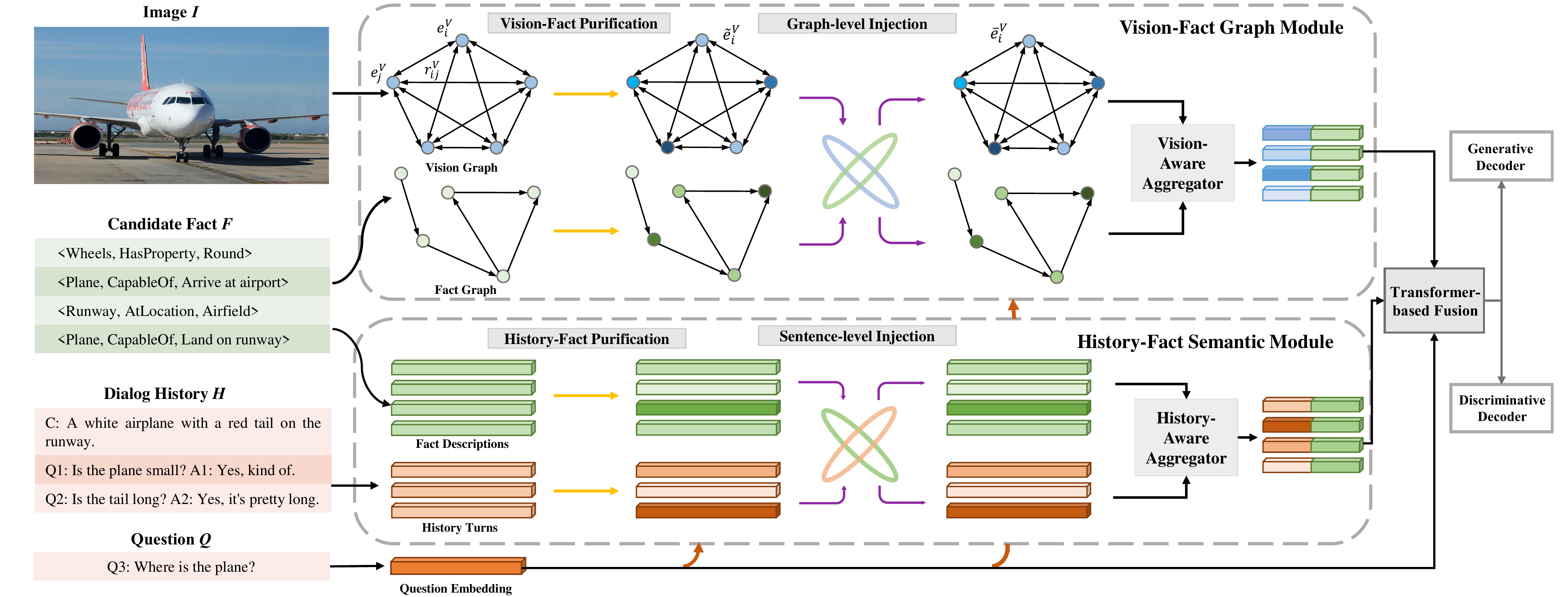}
\caption{Overview structure of RMK. The model mainly contains two modules: Vision-Fact Graph Module and History-Fact Semantic Module, both of which contain three operators: Purification (yellow arrows), Injection (pink arrows) and Aggregator. And the orange arrows in the figure denote the question-guided way. }

\label{img:model}
\end{figure*}

As shown in Fig.\ref{img:model}, our model consists of two modules: \emph{Vision-Fact Graph Module}, \emph{History-Fact Semantic Module}. Specifically, Vision-Fact Graph Module converts knowledge triplets to graph-level representation and further injects the commonsense knowledge into the graph-level vision bank. History-Fact Semantic Module involves sentence-level facts to the dialog history via cross-modal attention-based operations. Both two modules adopted three units, i.e. purification, injection, and aggregator to filter and incorporate relevant knowledge information. Finally, we adopt transformer-based multi-modal fusion and generate the response by the decoders. 

Our contributions can be summarized as follows: 

\begin{itemize}
\item We propose a novel method to represent commonsense knowledge in multi-structure: graph-level and sentence-level, to better suit the character of visual dialog and complement relevant information.
\item Furthermore, we adopt a multi-structure reasoning network to encode vision-fact graph knowledge and history-fact semantic knowledge, to extract implicit dependence in different modalities. The principled ablation study and visualization show how different modules work in our model. 
\item We conduct comprehensive experiments on two datasets: \emph{VisDial v1.0}~\cite{das2017visual} and \emph{VisDialCK}~\cite{agarwal2020history}. Note that VisDiaCK (a validation subset of VisDial v1.0) is a collection of commonsense-required questions in Visual Dialog. The results demonstrate the superiority our model.
\end{itemize}

\section{Related Work}

\noindent {\bf Visual Dialog.}
For the visual dialog task~\cite{das2017visual}, it aims to generate responses depending on an image, a caption, and the dialog history. LF~\cite{das2017visual}, MN~\cite{das2017visual}, CorefNMN~\cite{KotturSatwik2018Visual} and CoAtt~\cite{garderes2020conceptbert} utilize kinds of attention mechanisms as the backbone to locate the related visual objects. To solve the history-required problems such as visual co-reference, RVA~\cite{Niu2018Recursive} design recursive
visual attention, inferring the co-reference through recursively inspecting the history dialog and improving the visual attention.
Zheng \emph{et al.}~\cite{Zheng2019Reasoning} propose an EM-style inference algorithm to obtain the latent relations among history dialogs. MCA~\cite{agarwal2020history} focuses on an iterative question-conditioned context-aware graph, including both fine-grained visual and history semantics. 
DualVD~\cite{jiang2019dualvd} constructs a scene graph to represent the image, which emphasizes the essential role of vision for the referred visual content may change remarkably.
Another line of work targeted on response generation for visual dialog by carefully designed decoders. DMRM~\cite{chen2019dmrm} adopts multi-step reasoning based on dual attention to iteratively update related visual objects for a more relevant response. DAM~\cite{jiang2020dam} designs an adaptive decoder with memory to store the state of dialog history and visual information. Recently, pre-trained models~\cite{wang2020vdbert, murahari2020large} have also achieved impressive results in visual dialog. VisualBERT~\cite{murahari2020large} and VDBERT~\cite{wang2020vdbert} exploit large extra datasets to explore in visual dialog via pretraining language models. 

Though these works have achieved great success in performance, the commonsense-required problems are ignored and it still has space to improve by considering commonsense knowledge.

\noindent {\bf Knowledge-based VQA.}
Visual question answering (VQA)~\cite{Agrawal2017VQA} needs to give an accurate answer based on an image and a relevant question. Recently, there are many works proposed on knowledge-based VQA, including diverse benchmarks and systems. FVQA \cite{wang2017fvqa} is a fact-based VQA dataset that provides image-question-answer-supporting fact tuples. KBVQA \cite{wang2017explicit} divides data into three categories in which it needs visual concept, basic common sense, or higher-level knowledge with explicit reasoning. KVQA \cite{shah2019kvqa} consists of questions requiring world knowledge of named entities in images. Furthermore, OK-VQA \cite{marino2019ok} covers 11 categories of knowledge, such as cooking and food, science and technology, plants and animals, etc. 

Another line is the knowledge-based VQA models tapping into knowledge representations and reasoning strategies. 
Out of the Box \cite{narasimhan2018out} applies graph convolution networks to reason on the knowledge graph, whose nodes are attached by image and semantic embeddings. In addition, Mucko~\cite{zhu2020mucko} reasons on visual, fact, and semantic graphs separately, and utilizes cross-modal networks to aggregate information together for knowledge reasoning. KRISP \cite{marino2021krisp} employs a BERT-pretrained model to better understand semantics and exploit implicit knowledge. MAVEx~\cite{wu2021multi} votes among textual and visual knowledge from different sources.  
However, these works cannot apply to visual dialog directly, since visual dialog demands reasoning on both dialog history and image. Thus, how to design a knowledge fusion scheme adaptive to visual dialog appears particularly significant. Inspired by this, we design a multi-structure knowledge model to densely interact with both visual and dialog components in visual dialog. 

\noindent {\bf Vision and Language Modeling.} Approaches for multimodal vision and language tasks have explored diverse modeling strategies, such as GNN-based models (e.g.~\cite{jiang2019dualvd} ) or transformer-based ones (e.g.~\cite{wang2020vdbert}). Teney \emph{et al.} \cite{teney2018tips} propose the first GNN-based VQA method, which builds a scene graph of the image and parses the sentence structure of the question. Li \emph{et al.}~\cite{li2019relation} encodes each image into a graph and model inter-object relations via graph attention mechanism. Huang \emph{et al.} \cite{huang2020aligned} propose a novel dual-channel graph convolutional network to better integrate visual and textual information. GNN-based methods have also achieved impressive progress in visual dialog~\cite{jiang2019dualvd, chen-etal-2021-gog}, benefiting from the reasoning ability of graph network.

Over the past few years, multimodal transformers have made significant progress through pre-training on large-scale image and text pairs and then fine-tuning on downstream tasks. VisualBERT \cite{murahari2020large}, Unicoder-VL~\cite{li2020unicoder} and VL-BERT~\cite{su2019vl} propose the single-stream architecture on both images and text. ViLBERT~\cite{lu2019vilbert} and LXMERT \cite{tan2019lxmert} propose a two-stream architecture to process visual and textual information independently first and fused them later. CLIP~\cite{radford2021learning} aligns visual and language representations by contrastive learning and achieves state-of-the-art results in image-text retrieval.

Different from these work that uses transformer or other methods separately, our model first infers on the multi-structure knowledge with GNN's reasoning ability and then fuse different modalities via a transformer to better improve the interpretability and performance.

\section{Methodology}

The visual dialog tasks are as follows: given an image \emph{I} and the dialog history $H=\lbrace C,\left(Q_{1},A_{1} \right),...,\left(Q_{t-1},A_{t-1} \right)\rbrace$, where \emph{C} is the image caption. The task is to infer the best answer to the current question $Q_t$ by ranking a list of 100 candidate answers. Our work mainly focuses on the protocol of introducing external \textit{commonsense knowledge} to enhance the visual dialog system to reason for better answers. Based on the characteristics of the image and the dialog history, we observe commonsense knowledge as two profiles: graph-level and sentence-level. On top of them, we incorporate them into the dialog system adaptively, and we also visualize the reasoning clue in Fig.\ref{img:visualize}.

\subsection{Multi-structure Facts Representation}
\label{multi-facts-rep}
The image and dialog history are two key components in visual dialog. For the image, visual graph is widely adopted to handle the object relation~\cite{jiang2019dualvd}, and the dialog history is indispensable for its contextual information~\cite{Zheng2019Reasoning}. Therefore, single-structure commonsense knowledge cannot meet the diverse information demand. To fit the characteristics of them in a visual dialog, we represent commonsense knowledge in two aspects: \textit{sentence-level facts} and \textit{graph-level facts}. 

\noindent {\bf Sentence-level Facts.} In open-domain conversational systems, the semantics shared with commonsense knowledge is vital for establishing effective interactions \cite{zhou2018commonsense}. To capture the contextual semantics of fact triplets {\ttfamily <subject, relation, object>}, we convert it to semantic domain as the fact description ``{\ttfamily subject relation object}''. Then feed the description to an LSTM to get the sentence-level facts representation ${s}_{i}^F$.

\noindent {\bf Graph-level Facts.} The graph structure has great capability in gripping the relation between the entities. Thus, we utilize the graph structure to further underline the relationship between each commonsense knowledge entity complementary to visual graph. In detail, the graph-level facts are denoted as $G^F=(E^F,R^F)$, in which the node is fact entity $e_i^F \in E^F$. To enhance the semantic information in the fact graph, the edge $r_{ij}^F \in R^F$ can be calculated as:
\begin{equation}
    r_{ij}^F= tanh(\textbf{W}_r[{r}_{ij}^h,{r}_{ij}^d])
\end{equation}

where ${r}_{ij}^d$ is Fact Description representation corresponding to entity $e_i$ and $e_j$, ${r}_{ij}^h$ is the embedding of {\ttfamily relation} in the triplet. ``$[\cdot ,\cdot ]$'' denotes concatenation, and $\textbf{W}_{r}$ (as well as $\textbf{W}_{1}$,$\textbf{W}_2$, ..., $\textbf{W}_{n}$ mentioned below) are learned parameters in linear layers. 

To find the optimal supporting facts, we first retrieve relevant candidate facts from the knowledge base of facts~\cite{speer2017conceptnet}, following a score based approach proposed in \cite{narasimhan2018out}. We compute the cosine similarity of the embeddings of every word in the fact with the words in the caption and the words of visual concepts detected in the image. Then we average these values to assign a similarity score to the fact. These facts are sorted based on the similarity and the highest scoring facts are retained.

\subsection{Vision-Fact Graph Module}
\label{graph encoder}

For the objects in the image lacking relation information~\cite{jiang2019dualvd}, we combine the image with graph-level facts. As for the encoding strategy of image, we adopt the recent standard scheme~\cite{guo2020iterative}, conducting a graph for the image. This module mainly contains three units to filter and select informative vision and fact information: \textit{Vision-Fact Purification}, \textit{Graph-Level Injection} and \textit{Vision-Aware Aggragator}, shown in Fig.\ref{img:model}.

\textit{Vision-Fact Purification.} It aims to filter out less relevant information, for there may exist amounts of redundant information in the image and fact knowledge graph. In the visual feature graph $G^V=(E^V,R^V)$, the nodes $\bm{E}^V=\{e_i^V\}^{N}$ are visual entity features extracted by a detector, where $N$ is the number of detected objects. The edges $\bm{R}^V = \{r_{ij}^V\}^{N\times N}$ are the visual relations between nodes provided by a visual relationship encoder~\cite{zhang2019large}. The construction of the fact graph is described in Sec.\ref{multi-facts-rep}. Then we adopted relation-aware GCN~\cite{jiang2019dualvd} methods to aggregate relation information among the entities in the vision graph and fact graph. And it results to purified vision feature $\widetilde{E}^V$ and fact feature $\widetilde{E}^F$, respectively.

\begin{equation}
\label{eq:purification}
\begin{split} 
    \widetilde{E}^V =& GCN(E^V,R^V) \\
    \widetilde{E}^F =& GCN(E^F,R^F)
\end{split}
\end{equation}

\textit{Graph-Level Injection.}
The graph-level facts contain diverse knowledge, while the image may retain noisy entities that lack relevant information. The Graph-level Injection introduces external knowledge to help understand the visual information comprehensively, and also incorporates the visual knowledge into the facts graph to enhance the supported facts. 

It strengthens the image information with commonsense knowledge, while further grasping the most relevant facts guided by vision, through cross-graph interaction. Specifically, to equip the image with useful facts, the graph message $\widetilde{v}_i^M$ is transferred from facts $\widetilde{v}_j^F$ to visual entity $\widetilde{v}_i^V$ between two graphs. The facts-injected image entity $\bar{v}_i^V$ is generated as follows:

\begin{equation}
    \setlength\abovedisplayskip{-5pt}
    \setlength\belowdisplayskip{8pt}
\label{eq:injection}
\begin{split}
\gamma _{ij} \!= \!softmax &(\!\textbf{W}_{\gamma}  ( \tanh ( \!\textbf{W}_1 [Q_t,\widetilde{e}_i^V,\widetilde{e}_j^F])) \\
\widetilde{e}_i^M &= \sum_{j=1}^{N}\gamma_{ij} \widetilde{e}_j^F \\
\bar{e}_i^V =  &\tanh ( \textbf{W}_2 [\widetilde{e}_i^V, \widetilde{e}_i^M ])
\end{split}
\end{equation}

\noindent Where $Q_t$ is the question feature encoded by LSTM. We adopt additive attention~\cite{bahdanau2014neural} which is the concatenation followed by the weight matrix. The vision-injected facts entity $\bar{e}_i^F$ can be gained by swapping the position of $\widetilde{e}_j^F$ and $\widetilde{e}_i^V$ in the equations.

\textit{Vision-Aware Aggregation.} After Graph-Level Injection, the entities in a graph are injected with local complementary information from the other. We then aggregate facts graph to global representation via attention mechanism, and further concatenate it with visual features. The aggregated vision-fact representation $\bar{I}$ can be gained by:

\begin{equation}
\label{eq:aggregator}
\begin{split}
\delta_i = & softmax (\textbf{W}_{\delta}  (Q_t \circ ( \textbf{W}_3  \bar{e}_i^F)))\\[1ex]
\bar{I}_j & =  \textbf{W}_v [\widetilde{e}_j^V, \sum_{i=1}^{N}\delta_i \bar{e}_i^F]
\end{split}
\end{equation}


\subsection{History-Fact Semantic Module}

Distinct from the image, the dialog history has different characteristics in manifestations. The contextual relation information is included in the sentences implicitly, and the graph-level facts have limited ability in handling the semantics among sentences. Thus, we further introduce the sentence-level facts, which are denoted as ${\{s_i^F\}^{K}}$, where $K$ is the number of facts. The dialog history is denoted as $\{s_i^H\}^{T}$, where $T$ is the rounds of history. We adopted similar methods in previous graph module, after minor modification, to filter and fuse them: \textit{History-Fact Purification}, \textit{Sentence-level Injection} and \textit{History-Aware Aggregator}. 

In this module, \textit{History-Fact Purification} aims to evaluate the relevance of textual facts and history to the current question. Specifically, the sentence-level facts are purified by the guidance of question-aware attention.

\begin{equation}
\label{eq:encode}
\begin{split}
\eta _{i} = softmax &( \textbf{W}_{\eta}({Q}_{t}  \circ \textbf{W}_7  s_{i}^F))\\[1ex]
\widetilde{s}_{i}^H  = \ \ & \eta _{i}  s_{i}^F
\end{split}
\end{equation}

\noindent And the purified history features are gained in the same way.

As for \textit{Sentence-level Injection} and \textit{History-Aware Aggregator}, we similarly adopt the paradigm in Graph Module. And we computed Eq.\ref{eq:injection} and Eq.\ref{eq:aggregator} on the top of textual features, finally resulting to aggregated history-fact features $\bar{H}$. It can enrich dialog history and related facts with each other. 

\subsection{Multi-modal Fusion}
After obtaining the fact-aware representations, we fuse the question representation $Q_t$, history-fact feature $\bar{H}$, vision-fact feature $\bar{I}$ through a multi-modal fusion strategy. It can be any existing visual dialog model to learn the joint representation. In our experiments, we adopt a light-weight transformer-based method LTMI~\cite{nguyen2020efficient} to fuse them. 

\begin{equation}
    {E} =  \mathcal{F} (Q_{t},\bar{I}, \bar{H})
\end{equation}

Then the fused representation $E$ is fed to the decoder to generate responses to the given question. As for the decoder, we follow the previous studies~\cite{das2017visual} to set discriminative and generative decoders and adopt multi-task learning~\cite{nguyen2020efficient} by minimizing the sum of the generative loss and the discriminative loss.

\begin{table}[t] 
\centering
\caption{Result on VisDial v1.0 val set using generative decoder.}
\resizebox{0.49\textwidth}{!}{
\begin{tabular} {lcccccc}
\hline                       
Method &  NDCG$\uparrow$& {MRR$\uparrow$} & {R@1$\uparrow$} & {R@5$\uparrow$} & {R@10$\uparrow$} & {Mean$\downarrow$}  \\

\hline
MN \cite{das2017visual} &51.86 &47.99 &38.18 &57.54 &64.32 &18.60 \\
CoAtt \cite{wu2018areyou} &59.24 &49.64 &40.09 &59.37 &65.92 &17.86  \\
DMRM \cite{chen2019dmrm} & - &50.16 &40.15 &60.02 &67.21 &15.19\\
DAM \cite{jiang2020dam} &60.93 &50.51 &40.53 &60.84 &67.94 &16.65\\
KBGN \cite{jiang2020kbgn} & 60.42 &50.05 &40.40 &60.11 &66.82 &17.54\\
GoG \cite{chen-etal-2021-gog} & 62.63 &51.32 &41.25 &61.83 &69.44 &15.32\\
\hline
LTMI \cite{nguyen2020efficient}   & 61.61 &50.38 &40.30  & 60.72  & 68.44 & 15.73   \\
LTMI-RMK  &\textbf{63.57} & \textbf{51.76} & \textbf{41.56} & \textbf{62.16} & \textbf{69.83} & \textbf{15.05}  \\
\hline
\end{tabular}  
}
\label{table:gen}
\end{table}

\begin{table}[t] 
\centering
\caption{Results on VisDial v1.0 test-std set using discriminative decoder. Underline are the highest results except for pretraining-based models, which are trained with extra training data. }
\resizebox{0.49\textwidth}{!}{
\begin{tabular} {lcccccc}
\hline                       
Method & NDCG$\uparrow$ & MRR$\uparrow$ & R@1$\uparrow$ & R@5$\uparrow$ & R@10$\uparrow$ & Mean$\downarrow$  \\

\hline
LF \cite{das2017visual}  & 45.31 &55.42 & 40.95 & 72.45 & 82.83 & 5.95 \\
MN \cite{das2017visual} &47.50 & 55.49 & 40.98 & 72.30 & 83.30 & 5.92 \\  
CorefMN \cite{KotturSatwik2018Visual} &54.70 & 61.50 & 47.55 & 78.10 & 88.80 & 4.40 \\
RvA \cite{Niu2018Recursive} &55.59 &63.03&49.03&80.40&89.83&4.18\\
DualVD \cite{jiang2019dualvd}& 56.32 & 63.23 & 49.25 & 80.23 & 89.70 & 4.11 \\
CAG \cite{guo2020iterative} &56.64 & 63.49 & 49.85  &80.63 & 90.15 & 4.11 \\
KBGN \cite{jiang2020kbgn} &57.60 &{64.13} &50.47 &80.70 &90.16 &\textbf{4.08} \\
GoG \cite{chen-etal-2021-gog} &60.38 &63.13 &49.88 &79.65 &89.05 &4.39 \\

\hline
VDBERT \cite{wang2020vdbert} &\textbf{75.35} &51.17 &38.90 &62.82 &77.98 &6.69 \\
VisualBERT \cite{murahari2020large} &74.47 &50.74 &37.95 &64.13 &80.00 &6.28 \\
\hline
LTMI \cite{nguyen2020efficient}   &\underline{60.92} &60.65 &47.00 &77.03 &87.75 &4.90    \\
LTMI-RMK  &58.48 & \textbf{64.14} & \textbf{50.58} & \textbf{80.72} & \textbf{90.28} & 4.14  \\
\hline
\end{tabular}  
}
\label{table:dis}

\end{table}

\begin{table}[t] 
\centering
\caption{Results comparison on VisDialCK using discriminative decoder, where ${\dagger}$ means re-implemented with the same settings as ours for fair comparison.}
\resizebox{0.49\textwidth}{!}{
\begin{tabular} {lcccccc}
\hline   {Method} & {NDCG$\uparrow$} & {MRR$\uparrow$} & {R@1$\uparrow$} & {R@5$\uparrow$} & {R@10$\uparrow$} & {Mean$\downarrow$}  \\

\hline

${\rm LF}^{\dagger}$ \cite{das2017visual}  & 53.46 & 55.53 & 41.32  & 76.95  & 87.04  & 4.61   \\
${\rm MN}^{\dagger}$ \cite{das2017visual}  & 55.06 & 56.18 & 41.47  & 77.32   & 87.45 & 4.36  \\  
${\rm DualVD}$ \cite{jiang2019dualvd}  & 55.48 & 58.77  & 42.55 & 81.01   & 88.30   & 3.93   \\
\hline
LTMI \cite{nguyen2020efficient} & 58.74  & 58.12    & 43.78    & 80.27   & 88.23   & 4.02   \\
LTMI-RMK & \textbf{60.94} & \textbf{65.78} & \textbf{54.92} & \textbf{81.76} & \textbf{90.23} & \textbf{3.91} \\

\hline
\end{tabular}  
}
\label{table:sotaCK}
\end{table}

\section{Experiments}
\label{sec:experiments}

\subsection{Datasets} 

\noindent {\bf VisDial v1.0.}
For VisDial v1.0 dataset, the train, validation, and test splits contain 123k, 2k, and 8k dialogs, respectively. In “train” and “val”, each image is accompanied by a 10-round dialogue, while in “test”, each image is followed by random rounds of question-answer pairs and an ongoing question for answer prediction. The training split is composed of 123k images and each dialog consists of 10-round QA pairs for each image. The following metrics are adopted: mean reciprocal rank (MRR), recall@$k$ ($k$ =1, 5, 10), mean rank (Mean), and normalized discounted cumulative gain (NDCG). A lower value for Mean and higher for other metrics are desired. Note that we train the model on the VisDial v1.0 training set, and evaluate the model on the VisDial v1.0 val, test, and VisDialCK.

\vspace*{0.8\baselineskip}
\noindent {\bf VisDialCK.} 
For the purpose of verifying the effectiveness of RMK on commonsense-required questions in visual dialog, we also conduct evaluations on a commonsense-required dataset called {\em VisDialCK}. It is first proposed by \cite{agarwal2020history}, in which they conducted crowd-sourcing on VisDial v1.0 val to annotate the dialog into different categories, among which commonsense-required and history-required are the most two except for normal VQA kind (don't need history and commonsense). However, they only focus on history-required ones. So we further collect commonsense-required ones from their raw data to form VisDialCK, a subset of VisDial v1.0 val, which contains 940 history-required dialog rounds. It can properly reflect the model's capability to deal with the knowledge-required dialogs.

\subsection{Implementation Details} 

To build the vocabulary, we retain words in the dataset with word frequency greater than 5. Each word in the dialog is embedded into a 300-dim vector with the GloVe embedding initialization \cite{Pennington2014Glove}. The maximum sentence length of the dialog history and the current question are set to 20. The hidden state size of Transformer blocks is all set to 512. We adopt Adam optimizer with the initial learning rate of 4e-3 and final learning rate of 5e-5 via cosine annealing strategy with 16 epochs. The mini-batch size is 15 and the dropout \cite{srivastava2014dropout} ratio is 0.5. The model is trained with a multi-class N-pair loss. We choose the widely adopted ConceptNet as the external commonsense knowledge source \cite{speer2017conceptnet}. Following \cite{anderson2018bottom-up}, we use bottom-up features of 36 proposals from images using a Faster-RCNN \cite{Ren2017Faster} pre-trained on Visual Genome \cite{krishna2017visual} to get a bag of object-level 2048-d image representations. For the results on test set, we only report results for our best performing models as the number of allowed submissions to the challenge is limited.

\begin{figure*}[t]
\centering
\includegraphics[width=1.0\textwidth]{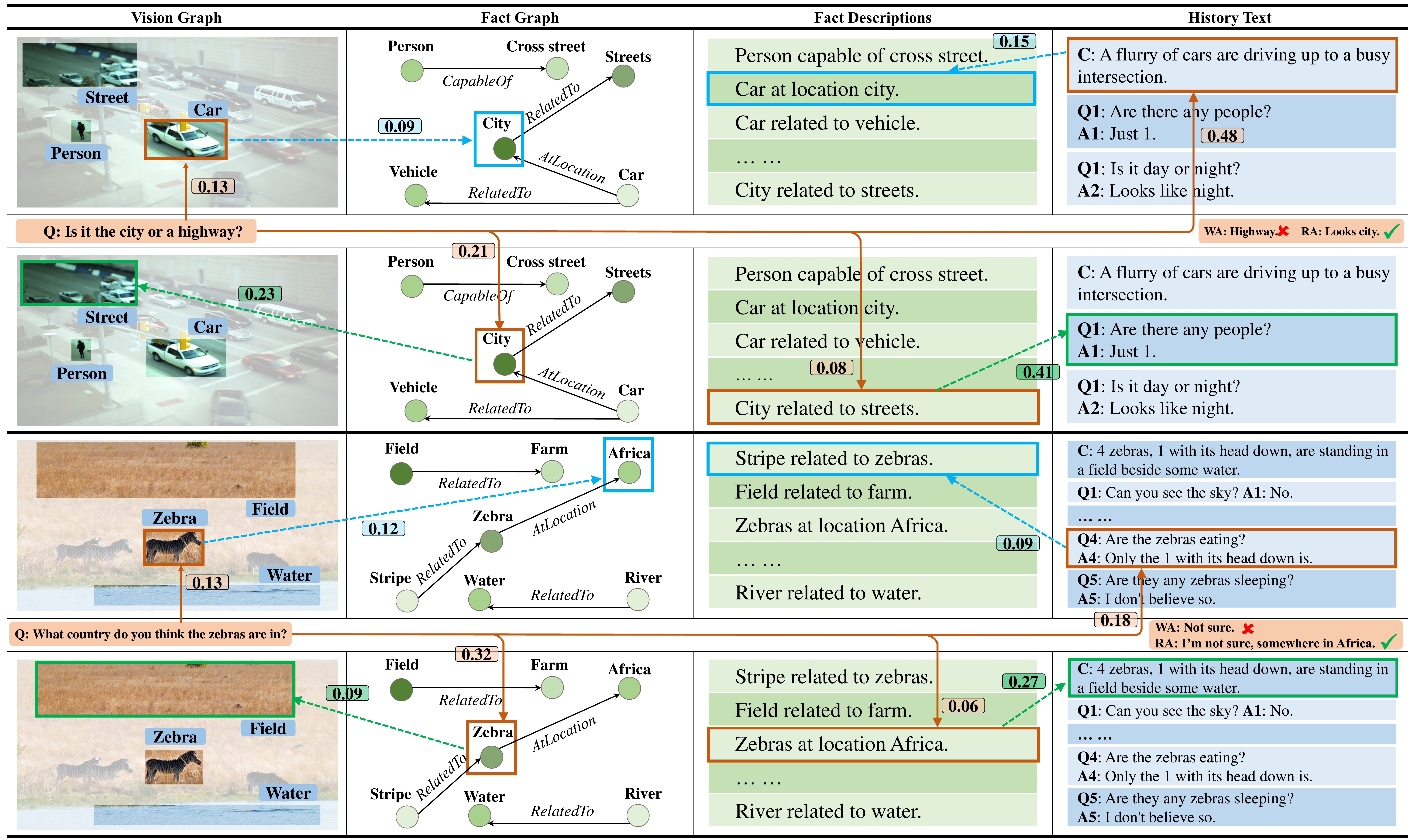}
\caption{Qualitative results from our RMK. The WA means the wrong answer predicted by \emph{LTMI}, while RA means the right answer by \emph{LTMI-RMK}. The decimals on the arrows from question Q to other modalities indicate the normalized question-guided attention. The arrows between facts and visual graph or history text denote the cross-modal interaction weight, displaying the complementary information.}

\label{img:visualize}
\end{figure*}

\begin{figure*}[t]
\centering
\includegraphics[width=0.98\textwidth]{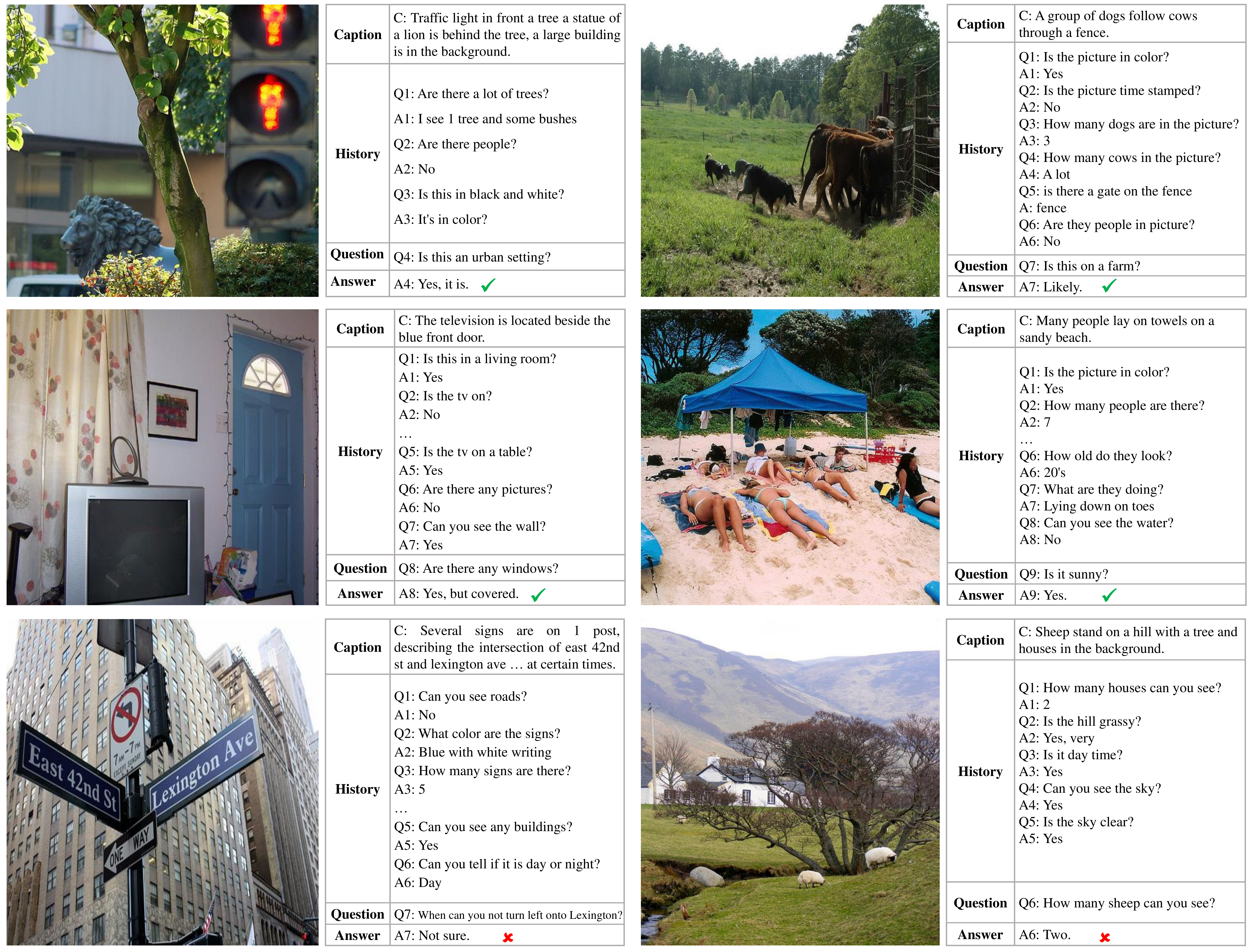}
\caption{More qualitative examples of our model. We show the caption, dialog history, question, image, and the answers generated by our proposed RMK model. }

\label{img:vis2}
\end{figure*}

\subsection{Comparison Results}

\noindent {\bf Baselines.} 
In our experiment, the compared methods mainly include: (1) Fusion-based and Attention-based models: LF~\cite{das2017visual}, MN~\cite{das2017visual}, CorefNMN~\cite{KotturSatwik2018Visual}, RvA~\cite{Niu2018Recursive}, DMRM~\cite{chen2019dmrm}, DAM~\cite{jiang2020dam}. (2) The pretraining model: VDBERT~\cite{wang2020vdbert} and VisualBERT~\cite{murahari2020large}. (3) Graph-based models: DualVD~\cite{jiang2019dualvd}, FGA~\cite{schwartz2019factor}, CAG\cite{guo2020iterative} , KBGN~\cite{jiang2020kbgn}. These methods are our mainly compared baselines.

\vspace*{0.3\baselineskip}
\noindent {\bf Generative Results.} 
First, we compare the performance of generative results of different models. As shown in Table \ref{table:gen}, our method outperforms all the compared methods with large margins on the val v1.0 split. Comparing with the results of LTMI~\cite{nguyen2020efficient} without commonsense knowledge, our model improves NDCG for 62.63 to 63.57 (+1.96), MRR from 50.38 to 51.76 (+1.38), R@1 from 40.30 to 41.56 (+1.26), Mean from 15.73 to 15.05 (+0.68) and more than 1\% on other metrics. 
Notice that GoG~\cite{chen-etal-2021-gog} additionally parses the words relations in a question and builds a more complex graph-over-graph network.
Our RMK validates that when incorporating commonsense knowledge, it improves significantly and outperforms other compared models on all metrics. It proves that RMK can improve the performance of visual dialog models by introducing explicit knowledge reasoning, which also illustrates that commonsense knowledge is helpful for visual dialog.

\begin{table}[t] 
\caption{Ablation study of model design on VisDial val v1.0. F in the second block is short for facts.}
\setlength{\belowcaptionskip}{-2mm} 
\centering
\resizebox{0.49\textwidth}{!}{
\begin{tabular}{lcccccc}
\hline
 Model & {MRR$\uparrow$} & {R@1$\uparrow$} & {R@5$\uparrow$} & {R@10$\uparrow$} & {Mean$\downarrow$}  & {NDCG$\uparrow$} \\ 
\hline
  LTMI        & 62.32 & 48.94 & 78.65 & 87.88 & 4.86  & 62.72 \\
  LTMI-RMK    & 65.08    & 51.78 & 81.62 & 90.48 & 3.98  & 60.68\\
 \hline
  w/o All F     & 63.92    & 50.32 & 80.13 & 89.27 & 4.43   & 58.37   \\
  w/o Sentence F   & 64.97  & 51.26 & 81.32 & 90.12 & 4.09  & 59.63    \\
  w/o Graph F    & 64.48  & 50.86 & 80.82 & 89.74 & 4.24  & 59.21   \\
\hline
  w/o Purificaition   & 64.84   & 50.92 & 80.73 & 90.23 & 4.13 & 59.12     \\
  w/o Injection     & 63.92  & 51.13 & 80.78 & 90.07 & 4.42  & 58.72   \\
  w/o Aggragator  & 64.16   & 50.75 & 80.91 & 89.83 & 4.20   & 58.63      \\
\hline

\end{tabular}}
\label{table:abl1}
\end{table}

\vspace*{0.3\baselineskip}
\noindent {\bf Discriminative Results.} 
We also compare discriminative results in Table \ref{table:dis}. Our method improves a lot compared to LTMI on the test-std v1.0 split, which is about +3\% on MRR, R@1, R@5, and R@10. Compared to previous non-pretrained models, our method also achieves significant improvement on most metrics, which proves that our method is effective and beneficial. The performance of our model even exceeds the performance of VDBERT~\cite{wang2020vdbert} on all the metrics except NDCG. Notice that the pretrain-based model(VDBERT and VisualBERT) works for they use a lot of extra train data except for VisDial train set. These observations show that RMK can assist in the improvement of visual dialog tasks. The reason why our method is effective is that we incorporate multi-structure of commonsense knowledge through our designed network.

\vspace*{0.3\baselineskip}
\noindent {\bf Results on VisDialCK.} 
To certify whether our model can deal with the commonsense-required questions successfully, we compare RMK with previous models on VisDialCK~\cite{agarwal2020history}. As shown in Table \ref{table:sotaCK}, RMK outperforms them on all metrics. Our model substantially improves a lot on LTMI, on MRR and R@1 by about +8\%, and on NDCG and R@10 by +2\%, which proves that the model can help with the questions that require commonsense. It verifies that the traditional methods can not answer the questions that require commonsense knowledge well. And the significant improvement also indicates that our method can indeed assist in handling the commonsense-required questions.

\begin{table}[t] 
\caption{Ablation study on different number of commonsense fact candidates on VisDial val v1.0. }
\centering
\resizebox{0.48\textwidth}{!}{
\begin{tabular}{ccccccc}
\hline
{ $\#$ facts}  & {MRR$\uparrow$} & {R@1$\uparrow$} & {R@5$\uparrow$} & {R@10$\uparrow$} & {Mean$\downarrow$}  & {NDCG$\uparrow$} \\ 
\hline
top 50        & 64.04    & 50.78    & 80.83    & 89.37      & 4.17   & 58.63      \\
top 100& \textbf{65.08} & \textbf{51.78} & \textbf{81.62} & \textbf{90.48} & \textbf{3.98}  & \textbf{60.68} \\
top 150     & 64.43      & 51.20    & 81.23   & 89.86       & 4.01       & 59.32   \\
top 200       & 64.65   & 51.32    & 81.17      & 90.23          & 4.05      & 58.95  \\
\hline
\end{tabular}}
\label{table:facts}
\end{table}

\subsection{Ablation Study}

In Table \ref{table:abl1}, we first remove the different levels of facts to validate the effect of multi-structure knowledge. The results in the second block show both the sentence-level and graph-level facts are crucial for visual dialog, and combining them can achieve better results.
In the second block, we investigate the importance of different operations in our model. w/o Purification removes the purification stage in both Vision-Fact Graph Module and History-Fact Semantic Module and others as the same. Without any of these three stages, the performance consistently drops, which validates the effectiveness of these adaptive strategies.

As shown in Table \ref{table:facts}, we vary the number of retrieved candidate facts for the model, in which top-k are ranked by the weighted score of fact confidence and visual object confidence. We achieve the best downstream metrics with the top 100 candidate facts (adopted by us). Fewer facts may not include the required facts for the questions, while too many facts may introduce much noise into the model.

\begin{table}[t]
    \centering
    \caption{Human evaluation on 100 sampled responses on VisDial val v1.0. M1: percentage of responses pass the Turing Test. M2: percentage of responses evaluated better than or equal to human responses.}
    \resizebox{0.30\textwidth}{!}{
    \begin{tabular}{l|cc}
    \toprule
     & \multicolumn{1}{c}{LTMI\cite{nguyen2020efficient}} & \multicolumn{1}{c}{RMK} \\ 
    \hline
    Method 1 (M1) & 54          & 65                       \\
    Method 1 (M2) & 62          & 68                       \\
    \bottomrule
    \end{tabular}
    }
    \label{tab:human}
\end{table}

\subsection{Human Study}
As shown in Table~\ref{tab:human}, we conduct human study to further demonstrate the effectiveness of our proposed RMK model. Our model achieves the highest scores both on the metrics M1 and M2 compared with LTMI model. These results show that our model can generate a contextually coherent response, which is more in line with human commonsense.

\subsection{Qualitative Results}
To figure out how the RMK model works, we visualize the reasoning paths on top of the multi-structure commonsense knowledge with vision and history information. Figure \ref{img:visualize} shows two examples, in which the first one comes from VisDialCK and the second comes from VisDial val set. There are two reasoning clues for answering the question: one is reasoning through vision or history to support facts (the row above questions in Fig.\ref{img:visualize}), and the other reasons from question directly to facts incorporated with vision or history information (the row below questions).

Take the first example for detailed analysis. When answering the given question ``\emph{Is it the city or a highway?}'', to determine what is the image about, the model focuses on the main object \emph{Car} which is directed to \emph{City} in Fact Graph. Similarly, reasoning from question through caption $C$ in history also leads to ``\emph{Car at location City}'' in Fact Descriptions. Moreover, as seen in the blocks below the question, the model can link the question directly to the relevant fact entity \emph{City} and fact description ``\emph{City related to streets}''. Finally, our model generates a more reliable answer ``\emph{Looks city}'' rather than ``\emph{Highway}'', which is more in line with commonsense compared to the one without facts knowledge.  
Similar observation exists in the second example. Faced with the difficult question of where the zebras are, RMK points the relevance of Africa in the facts and then chooses the optimal answer. With the commonsense knowledge, it generates a more informative answer ``\emph{somewhere in Africa}'' instead of a safe response ``\emph{Not sure}''. It illustrates that our multi-structure knowledge reasoning architecture can not only extract the required information from the facts, but also capture the underlying dependence from vision and history.

In addition, we supply more qualitative examples from our model as shown in Figure~\ref{img:vis2}. In the first four examples, our model can handle the diverse kinds of questions in visual dialog. The last two examples are the failure cases for our model. The second last one needs looking into the text on the image while our model not. For the last example, there are actually three sheep in the image, but the answer is “Two”. It shows that our model cannot well handle the question related to the text on the image (may need OCR as in TextVQA~\cite{singh2019towards}) and the complicated counting problem, which also remain open questions in multimodal systems.

\section{Conclusion}
In this paper, we introduce a novel model RMK for reasoning with commonsense knowledge in visual dialog. To properly suit the characteristics of dialog history and image in the task, we first represent commonsense knowledge at multi-structure level: sentence-level facts and graph-level facts. Then it captures and fuses relevant knowledge into visual dialog system, complementing with the visual graph and the history sentences. 
Experimental results on two datasets illustrate the superiority of our proposed model, and show the significant increase with external knowledge for VisDial task. The work will inspire research on visual dialog involving knowledge-based reasoning.

{\small
\bibliographystyle{ieee_fullname}
\bibliography{egbib}
}

\end{document}